\pdfoutput=1

\documentclass[11pt]{article}

\usepackage[final]{acl}

\usepackage{times}
\usepackage{latexsym}
\usepackage{multirow}  
\usepackage{hyperref}
\usepackage{tabularx, ragged2e}
\usepackage{makecell}
\usepackage{booktabs}
\usepackage{amssymb}
\usepackage{amsmath}
\usepackage{afterpage}
\usepackage{array}
\usepackage{cleveref}

\usepackage[T1]{fontenc}

\usepackage[utf8]{inputenc}

\usepackage{microtype}
\usepackage{todonotes}
\usepackage{inconsolata}
\usepackage{geometry}
\usepackage{graphicx}
\usepackage{float}
\usepackage{booktabs}

\title{Learning to Look at the Other Side: A Semantic Probing Study \\ of Word Embeddings in LLMs with Enabled Bidirectional Attention}

 \author{Zhaoxin Feng \and Jianfei Ma \and Emmanuele Chersoni \\ \and {\bf Xiaojing Zhao} \and {\bf Xiaoyi Bao} \\
         Language Science and Technology, The Hong Kong Polytechnic University\\
        \texttt{\{zhaoxinbetty.feng,jian-fei.ma,xiaojing.zhao,xiaoyi.bao\}@connect.polyu.hk,} \\ \texttt{emmanuele.chersoni@polyu.edu.hk}}

\begin{document}
\maketitle
\begin{abstract}{
Autoregressive Large Language Models (LLMs) demonstrate exceptional performance in language understanding and generation.  However, their application in text embedding tasks has been relatively slow, along with the analysis of their semantic representation in probing tasks, due to the constraints of the unidirectional attention mechanism. 

This paper aims to explore whether such constraints can be overcome by enabling bidirectional attention in LLMs. We tested different variants of the Llama architecture through additional training steps, progressively enabling bidirectional attention and unsupervised/supervised contrastive learning.



Our results show that bidirectional attention improves the LLMs' ability to represent subsequent context but weakens their utilization of preceding context, while contrastive learning training can help to maintain both abilities\footnote{Our code and data are released at: \url{https://github.com/Zhaoxin-Feng/semantic-probing-2025}. }}. 

\end{abstract}



\section{Introduction}

\begin{figure}[t]
\centering
\definecolor{pink_fillColor}{HTML}{fcf1f0}
\definecolor{pink_borderColor}{HTML}{fccccb}
\definecolor{yellow_fillColor}{HTML}{f3deb7}
\definecolor{yellow_borderColor}{HTML}{f9d580}
\definecolor{blue_fillColor}{HTML}{C4E1F6}
\definecolor{blue_borderColor}{HTML}{81BFDA}

\begin{tikzpicture}[scale=0.45,
    box/.style={draw, rounded corners=2pt, minimum height=6mm, minimum width=12mm},
    pinkbox/.style={box, fill=pink_fillColor,
    draw=pink_borderColor},
    yellowbox/.style={box, fill=yellow_fillColor,
    draw=yellow_borderColor},
    bluebox/.style={box, fill=blue_fillColor,
    draw=blue_borderColor},
    arrow/.style={->, bend left=20, thick},
    bidir/.style={<-, thick},
    label/.style={text=gray, font=\small},
    label_black/.style={text=black, font=\small}
]

\node[pinkbox] (t1) at (-1,6) {The};
\node[yellowbox] (b1) at (2,6) {bank};
\node[bluebox] (r1) at (6,6) {of the river};
\draw[arrow, yellow_borderColor, line width=1.1pt] (t1.north) to (b1.north);
\draw[arrow, yellow_borderColor, line width=1.1pt] (b1.north) to (r1.north);
\node[label] at (3.5,4.7) {Next token prediction in Llama};

\node[pinkbox] (t2) at (-1,2.7) {The};
\node[yellowbox] (b2) at (2,2.7) {bank};
\node[bluebox] (r2) at (6,2.7) {of the river};
\draw[bidir, blue_borderColor, line width=1.1pt] (t2.north) to[bend left=20] (b2.north);
\draw[bidir,blue_borderColor, line width=1.1pt] (b2.north) to[bend left=20] (r2.north);
\draw[bidir, blue_borderColor, line width=1.1pt] (t2.north) to[bend left=20] (r2.north);
\node[label] at (3.5,1.4) {Word representation in Llama};

\draw[blue_borderColor, dashed, line width=0.7pt] (-7,0.7) -- (8,0.7);

\node[pinkbox] (t3) at (-6,-1) {The};
\node[yellowbox,label_black] (m3) at (-2.5 ,-1) {[MASK]};
\node[bluebox] (r3) at (1.5,-1) {of the river};
\draw[arrow, yellow_borderColor, line width=1.1pt] (t3.north) to (m3.north);
\draw[bidir, yellow_borderColor, line width=1.1pt] (m3.north) to[bend left=20] (r3.north);
\node[label] at (-2,-2.3) {Masked language modeling in BERT};

\node[pinkbox] (t4) at (-6,-4.3) {The};
\node[yellowbox] (b4) at (-2.5,-4.3) {bank};
\node[bluebox] (r4) at (1.5,-4.3) {of the river};
\draw[bidir, blue_borderColor, line width=1.1pt] (t4.north) to[bend left=20] (b4.north);
\draw[arrow, blue_borderColor, line width=1.1pt] (b4.north) to[bend left=20] (r4.north);
\node[label] at (-2,-5.6) {Word representation in BERT};

\node[label_black, anchor=east] at (-2.5,4.5) {(a) Unidirectional};
\node[label_black, anchor=east] at (-3.3,3.7) {Attention};
\node[label_black, anchor=west] at (3.5,-2.5) {(b) Bidirectional};
\node[label_black, anchor=west] at (4.6,-3.3) {Attention};
\end{tikzpicture}

\caption{Comparison of attention mechanisms in Llama and BERT models. (a) shows Llama's unidirectional attention where prediction (orange arrows) and word representation (blue arrows) can only access one side context; (b) shows BERT's bidirectional attention where masked language modeling allows word representation to access both previous and subsequent context. }

\label{fig:attention_comparison}
\end{figure}
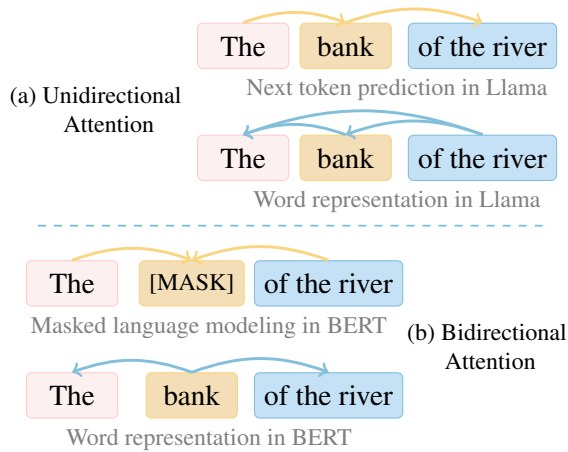

Decoder-only LLMs using autoregressive pretraining have achieved superior performance across language understanding and generation tasks, causing a major shift from the previous pretraining-then-finetuning paradigm dominated by encoder-only models~\citep{naveed2023comprehensive}. However, the community has been relatively slow in adopting them for word, sentence, and document embedding tasks because of their apparent limitations as text encoders, which have been speculated to be due to the lack of bidirectional attention~\citep{qorib2024decoder,behnamghader2024llm2vec, springer2024repetition}. As illustrated in Figure~\ref{fig:attention_comparison}, decoder-only LLMs can only access preceding contextual information during inference, resulting in word representations that encode information from the previous context, instead of the entire input sequence. 

This architectural constraint is potentially very limiting in tasks requiring fine-grained modulation of word meanings: while contextualized embeddings from encoder-only models marked significant progress compared to previous generation distributional models \citep{bommasani2020interpreting,chronis2020bishop}, the availability of the right-hand context might be important to capture subtle meaning nuances, and disambiguate the senses of polysemous words \citep{zhu-etal-2024-large,qorib2024decoder}.

Can this limitation be addressed? After all, the decoder-only architecture enables more efficient learning from all input tokens during pre-training, significantly improving sample efficiency compared to encoder-only counterparts~\citep{clark2020electra}, and this would be an important advantage if LLMs representations could be adapted to perform better in embedding tasks.


Given the above-mentioned research background, we propose the following research question: \textit{does bidirectional attention in LLMs enhance the quality of word meaning representations in LLMs? Could they achieve the same quality of embeddings extracted from encoder-only models?}

In this paper, we propose a probing study of different types of LLMs architectures on a pool of \textit{lexical semantic tasks}. Drawing inspiration from recent work on enabling bidirectional attention in autoregressive models~\citep{behnamghader2024llm2vec}, we compare the performance of Llama embeddings under the following configurations:
i) a base Llama architecture,
ii) the architecture in i) after an additional training step to enable bidirectional attention;
iii) the architecture in ii), after applying unsupervised/supervised contrastive learning.


Perhaps surprisingly, we found that bidirectional attention in itself does not improve the performance of Llama embeddings on our tasks: while it improves LLMs' ability to represent the right-hand context of a target word, it also seems to weaken the representation of the left context. Contrastive learning techniques often help the models to maintain both abilities, with Llama architectures getting on par or even outperforming bidirectional, BERT-based baselines on our tasks. Interestingly, we also found that adding bidirectional attention alone exacerbates the \textit{anisotropy}~\cite{ethayarajh2019contextual, cai2021isotropy, godey2024anisotropy} (a condition in which all vectors occupy just a narrow cone in the vector space) in all layers, resulting on average in higher similarity scores between the vectors of randomly sampled words. These findings reveal the potential of decoder-only LLMs in word embedding tasks,  offering insights into enhancing LLMs' representations with bidirectional attention and contrastive learning.

\section{Related Work}

\subsection{Representations of Word Semantics}

The representations of word semantics in NLP have undergone a remarkable development in the last two decades. Early methods like distributional semantic models (DSMs) derive semantic representations from statistical patterns of word co-occurrences in large text corpora, assuming that words with similar contexts have similar meanings~\citep{harris1954distributional,schutze1992dimensions,bullinaria2012extracting}. Later more efficient methods like Word2Vec and GloVe \citep{mikolov2013efficientestimationwordrepresentations,pennington2014glove} emerged, using neural networks to train word embedding representations more compactly, without time-consuming high dimensional spare data processing and high perplexity algorithms calculation~\cite{pennington2014glove,mikolov2013efficientestimationwordrepresentations}. 

However, these static vector models struggled with polysemy, as they assigned a single vector to each word regardless of context~\citep{faruqui-etal-2016-problems,gladkova-drozd-2016-intrinsic,wang2020static}. This limitation was addressed by contextualized embedding models such as ELMo and BERT~\citep{peters-etal-2018-deep,devlin2018bert}, in which word vectors are learned as a function of the internal states of the network, such that a word in different sentence contexts determines different activation states and is represented by a distinct vector~\citep{chersoni2021decoding}. 

Despite the advantages of representing context-specific meanings, contextualized vectors were shown to have a high level of anisotropy, i.e. they occupy just a narrow cone in the vector space, with the consequence that randomly-sampled words might also get high similarity values~\citep{ethayarajh2019contextual} and postprocessing techniques need to be applied to adjust the similarity metrics for anisotropy~\citep{timkey2021all}.

\subsection{Probing Linguistic Features in LLMs}
Probing-based methods for analyzing linguistic features have become prevalent for understanding the internal linguistic knowledge of language models~\citep{linzen2016assessing,hewitt2019designing,liu2019linguistic,wu2020perturbed,chersoni2021decoding,kauf2023event,matthews2024semantics,wang2024probing,liu2024fantastic}. The core idea involves using a simple diagnostic model (the “probe”, usually a linear classifier) to predict specific linguistic properties (e.g. animacy) from language model's output representations. If the model succeeds, we can infer that the representations of the language model encode that linguistic knowledge~\citep{chersoni2021decoding}. For instance, \citet{hewitt2019structural} presents a linear classifier to predict a target syntactic structure based on contextualized word representations to measure the syntactic information encoded in language models. 

For LLMs, the most prevalent method to probe the linguistic knowledge is to use the representation is the hidden state of the last layer~\citep{neelakantan2022text,behnamghader2024llm2vec,springer2024repetition,lee2025nvembedimprovedtechniquestraining}. We went for the last layer embedding also for an issue of methodological alignment with these approaches. Also, considering that the last hidden layer is processed through all model layers, in theory it contains the richest and most comprehensive context information, and thus our method employs these final layer hidden states to derive contextual representations of target words.



\subsection{Autoregressive LLMs as Text Encoders}

While ELMo and BERT's bidirectional attention gives them access to both the left and the right context around the target word, autoregressive LLMs like the GPTs~\citep{radford2019language,brown2020language} only use the context that comes before the target. Therefore, autoregressive LLMs are often considered sub-optimal for text embedding tasks. 

However, recent studies have shown that even under the constraints of
causal masking, LLMs are still capable of capturing certain contextual relationships~\citep{muennighoff2022sgpt,wang2023improving,behnamghader2024llm2vec,springer2024repetition}.  Among these studies, both~\citet{springer2024repetition} and~\citet{behnamghader2024llm2vec} specifically hypothesize that the limitation of LLMs lies in unidirectional attention: the former proposed “echo embeddings”, a method that feeds a target input sequence twice to a model to allow for the encoding of the context after the target; while the latter proposes an additional training step, called \textit{masked next token prediction}, to enable bidirectional attention in autoregressive models, and introduces further refinements based on unsupervised and supervised contrastive learning techniques, in order to improve the performance in sentence-level tasks. We ground our work in \citet{behnamghader2024llm2vec}'s LLM2Vec framework, using its publicly available models to observe how different training strategies affect the models' behavior in lexical semantic tasks.

\section{Experimental Setup}

\subsection{Model Selection}

Our experiments focused on Llama architectures \citep{touvron2023llama}, and particularly on Sheared-Llama \citep{xia2023sheared} and Llama 2 \citep{touvron2023llama2}. The former was chosen because it is a structurally-pruned and space-efficient version of the original model (we used the 1.3B version), and it is the closest in terms of parameter size to the most commonly-used bidirectional models (e.g. BERT and RoBERTa). We selected the 7B version of Llama 2 mainly to check if the trends identified with Sheared-Llama-1.3B were consistently observed in a bigger model.

Besides the base models (i.e. \textbf{Sheared-Llama-1.3B} and \textbf{Llama2-7B}), for our experiments we tested their variants augmented with the additional training steps of the LLM2Vec framework \citep{behnamghader2024llm2vec}: 1) the \textbf{Bi+MNTP} models underwent an additional training via \textit{masked next token prediction}, in which part of the input tokens was masked and the model had to reconstruct them on the basis of left and right context. For the prediction of each masked token, only the logits obtained from the previous token positions were used for computing the loss; 2) the architecture in 1), but with the addition of unsupervised  (\textbf{SimCSE}) and \textbf{Supervised} contrastive learning \citep{gao2022simcsesimplecontrastivelearning} on top of the bidirectional training. 
\citet{behnamghader2024llm2vec} added this step claiming that an autoregressive LLM with Bi+MNTP could be adequate for word-level tasks (they indeed obtain improved performance on standard benchmarks for POS Tagging and Named Entity Recognition), but contrastive learning is helpful to make their sequence representations a good fit  for sentence-level tasks as well.
All the augmented models are based on the same Sheared-Llama and Llama 2 architectures, and this gives us the chance to directly compare the effects of each additional training step. 

We also used the embeddings from BERT Base and BERT Large \citep{devlin2018bert} as our bidirectional baselines. For more details about our LLMs and the settings of the probe classifiers, the reader can refer to Appendix \ref{sec:appendix_model}.

\begin{figure*}[htbp]
    \centering
    \includegraphics[width=1.0\textwidth]{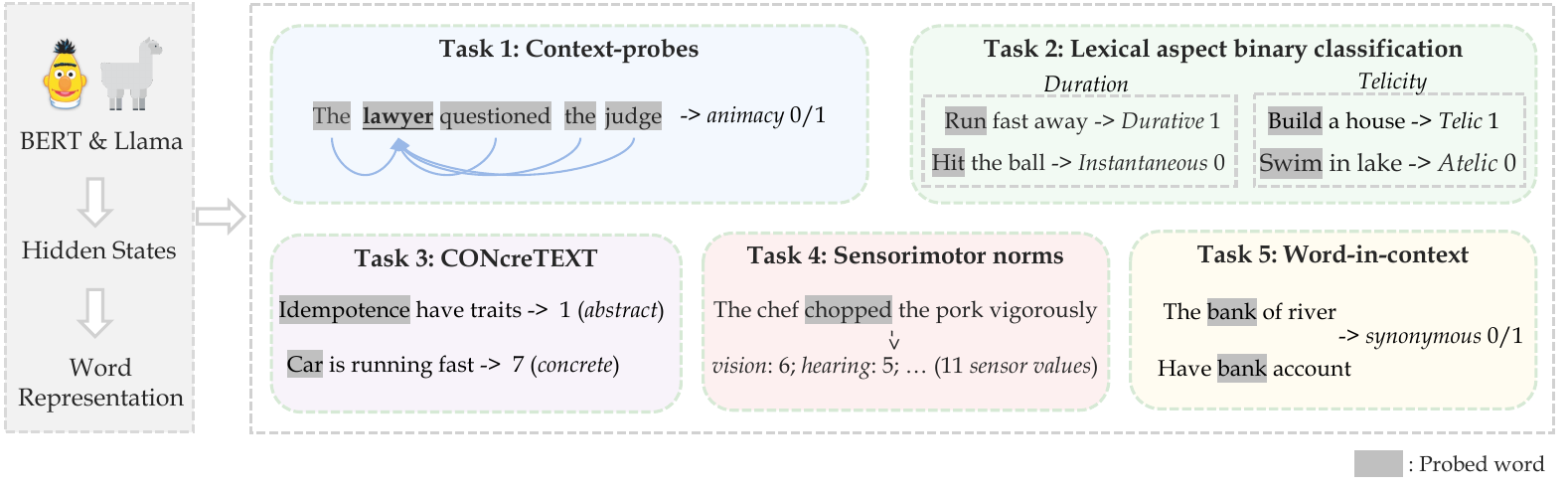} 
    \caption{Summary of five semantic probing tasks in our study: Tasks 1, 2, and 5 are classification tasks, with 0 and 1 denoting binary labels; Tasks 3 and 4 are regression tasks, where numbers (eg. 5, 7) indicate continuous values for the target variables. The “probed word” (highlighted) refers to the word whose contextualized representation is extracted for the probing task.}
    \label{fig:framework}
\end{figure*}


\subsection{Extracting Word Representations}
 

We leverage the hidden states from the final layer of LLMs to obtain contextualized representations of the target word for each lexical semantic task. Given a word $w$ within a sequence $c$, we first extract sequence representations as follows:

\begin{equation}
  \label{eq:hidden_states}
  \mathbf{H} = [\mathbf{h}_1, \mathbf{h}_2, ..., \mathbf{h}_n] \in \mathbb{R}^{n \times d}
\end{equation}
where $n$ denotes the sequence length (number of tokens) and $d$ represents the no. of hidden dimensions (e.g., 2048 in Sheared-Llama-1.3B).

For a word $w$ tokenized into $k$ subwords $\{t_1, t_2, ..., t_k\}$, let $\mathcal{I} = \{i_1, i_2, ..., i_k\}$ denote their positional indices in $\mathbf{H}$. The final word representation $\mathbf{v}_w$ is obtained through:

\begin{equation}
  \label{eq:mean_pooling}
  \mathbf{v}_w = \frac{1}{|\mathcal{I}|} \sum_{j \in \mathcal{I}} \mathbf{h}_j
\end{equation}

When $|\mathcal{I}| = 1$ (a single-token word), Equation~\ref{eq:mean_pooling} simplifies to $\mathbf{v}_w = \mathbf{h}_i$. This formulation ensures consistent representation for both single- and multi-token words while preserving contextual information from the final layer.


The first four tasks all involve a single target word in a sentence context, so we directly feed the target word's representation to a classifier or regression model for prediction. 
The exception was the \textbf{\textit{Word-in-Context}} (see Section \ref{sec:task}), which required comparing the meaning of a target word in two different sentences. In this task, we test three types of embeddings-derived features as input to a classifier: their \textit{cosine similarity}, the \textit{absolute values of their element-wise difference}, and their \textit{concatenation}. 

\subsection{Datasets and Tasks}
\label{sec:task}

\begin{figure*}[t!]
    \centering
    \includegraphics[width=1.0\textwidth]{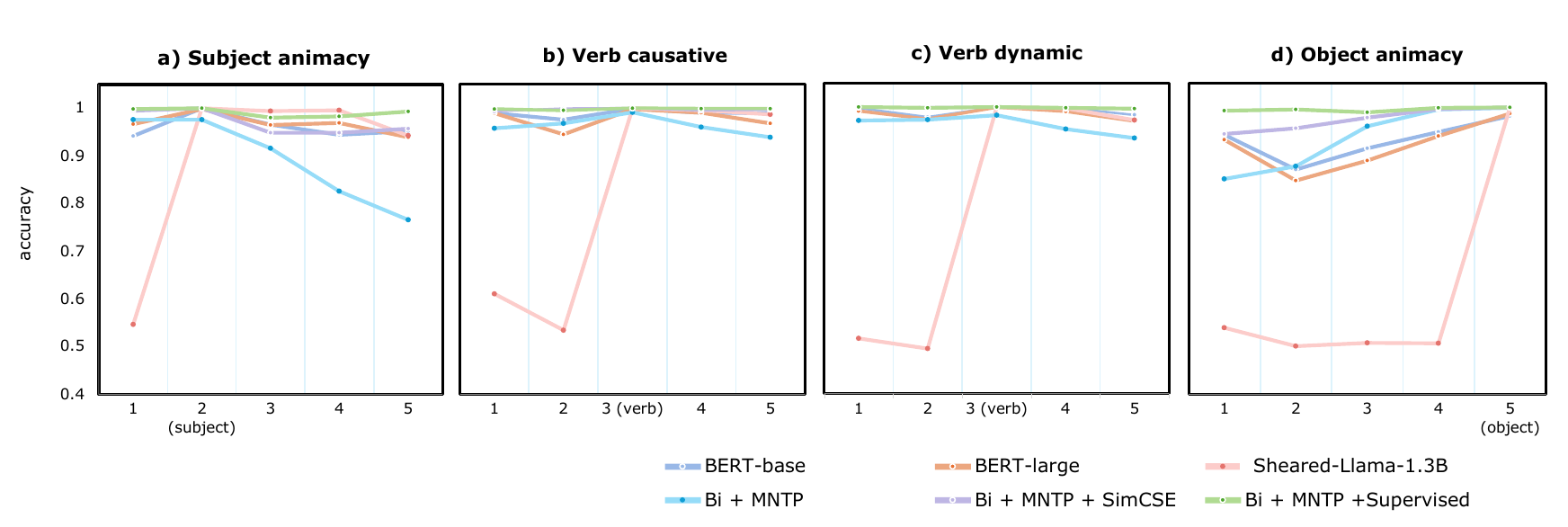} 
    \caption{Results of predicting subject animacy, verb causative/dynamic, and object animacy using each word in a sentence as probed words, scores extracted with Sheared-Llama-1.3B and its variants. The horizontal axis represents word indices in sentences (all with identical five-word syntactic structures).}
    \label{fig:context_probes}
\end{figure*}

Our study adopts five probing datasets (Figure \ref{fig:framework}). First, the \textit{\textbf{context-probes}} dataset by \citet{klafka-ettinger-2020-spying}, 
which is composed of sets of five-word sentences with a subject-verb-object structure and a binary property annotated for the verb or one of the nouns in the sentence, e.g. animate vs. inanimate for one of the nouns (the subject or the object); dynamic vs. static, and supporting causative-inchoative alternation for verbs. This is an easy task, targeting relatively stable properties of the target words regardless of context, but it can be useful to test the attention patterns of a model: in theory, LLMs with bidirectional attention should be able to solve it \textit{regardless of which token embedding is fed to the classifier} (i.e., even if the target word follows the token embedding, a contextualized embedding from a BERT model should still contain information from the right-hand context), while autoregressive LLMs should struggle more when the input embedding is extracted from a token preceding the target. 

A second task, targeting verbs, is \textit{\textbf{lexical aspect classification}} \citep{metheniti-etal-2022-time}. This dataset contains annotations about telicity and duration for verbs in a sentence context: given a verb embedding, a probing model has to determine whether it is telic/atelic or durative/stative. 
 


Next, we ran two regression tasks using two datasets targeting the semantics of words in context:
\textit{\textbf{CONcreTEXT}}~\citep{gregori2020concretext} and \textbf{\textit{contextualized sensorimotor norms}}~\citep{trott2022contextualizedsensorimotornormsmultidimensional}. The former contains annotations about the concreteness of a given word, with mean human ratings on a Likert scale ranging from 1 (totally abstract) to 7 (totally concrete); the latter contains mean human ratings for both verbs and nouns on 11 different sensorimotor domains, such as vision, hearing, etc. For each word, a sensorimotor rating indicates to what extent the corresponding sense is relevant to experiencing the concept referred by the target word in that specific context. In both cases, the goal for a probe model is to predict human ratings using as input the embedding of the target word \citep{fagarasan2015distributional,utsumi2018neurobiologically,thompson2018automatic,li2019mapping,chersoni2020automatic,chersoni2021decoding,flor2024three}.


For the final task, we employ the \textbf{\textit{Word-in-Context}}~\citep{pilehvar-camacho-collados-2019-wic} dataset. In each instance, a target word appear in two different sentences, and the probe model has to determine whether the word is being used in the same sense or not. 
Dataset details are described in Appendix~\ref{sec:appendix_data}. 


We selected the probing tasks to address different types of semantic features for different parts-of-speech (e.g. nouns and verbs), and the tasks demand a different level of contextual sensitivity: while the features of \textbf{\textit{context-probes}} or verb aspect should be stable for a target word across linguistic contexts, the regression tasks and the \textbf{\textit{Word-in-Context}}  were explicitly designed to require a deeper understanding of contextual meaning.

As for the probes, we tested both a linear and a non-linear model on top of the embeddings: the former was logistic regression for classification tasks and linear regression for concreteness and norms predictions; the latter was a multilayer perceptron (see Appendix~\ref{sec:appendix_result}).
In the main text, we only report the results of the Multilayer Perceptron, which achieved higher scores, while the results for the linear models are in  Appendix \ref{sec:appendix_result_lr}. All tasks were implemented on a single 40GB NVIDIA A100 GPU.

To ensure that our probing methodology actually evaluates the quality of contextual information in embeddings and to assess our selected tasks' sensitivity to context, we conducted two complementary experiments documented in the Appendix \footnote{The experiments were added upon a reviewer's request.}: 1) control tasks with randomly sampled labels (Appendix~\ref{sec:effective_probing}), and 2) out-of-context evaluations where models processed probed words in isolation rather than complete sentences (Appendix~\ref{sec:non_context}).

\section{Results and Analysis}

The experimental results yield three main findings derived from Tasks 1 and 2, Tasks 3 and 4, and the final task, respectively.

\textit{\textbf{Finding 1}: Bidirectional attention improves the LLMs' ability to represent subsequent context, but it weakens the utilization of the previous context. Contrastive learning techniques mitigate this trade-off by enhancing the model's ability to balance contextual understanding in both directions.}

\begin{figure}[!ht]
\includegraphics[width=0.45\textwidth]{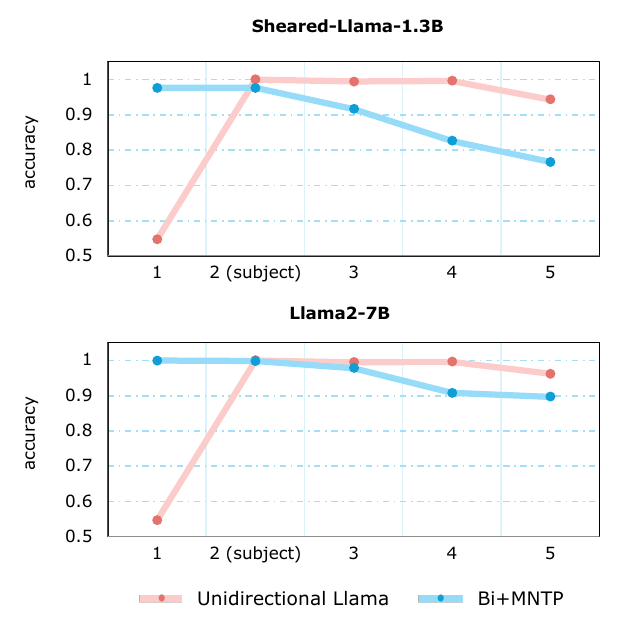} 
    \caption{Results of subject animacy subtask in Task 1 by comparing Sheared-Llama-1.3B to Llama2-7B.}
    \label{fig:wic2}
\end{figure}

 Figure~\ref{fig:context_probes} shows the results for the \textbf{\textit{context-probes}} dataset, for the base, unidirectional Llama, when the embedding of the probed word comes before the target word, the model has severe difficulty in correctly predicting the properties of the target word. 
 On the other hand, it can be seen that the performance of the bidirectional baselines is consistently high, regardless of the token embedding.
 Once bidirectional attention is enabled, however, the score pattern for Llama Bi + MNTP is more aligned with the BERT models.

There is, though, an important difference: by comparing Bi + MNTP and unidirectional Llama in Figure~\ref{fig:context_probes} a), we can find that the pink line shows relatively stable accuracy in the latter half, while the blue line exhibits a noticeable decline as the distance between target word and token of the input embedding increases. This indicates that enabling bidirectional attention may also weaken Llama's inherent ability to “see” the preceding context. 

At the same time, if we look at the models enhanced with contrastive learning (purple and green lines), we can notice that the additional training helps Llama to maintain representation quality in all the token positions. The results of the first task thus suggest that the additional training to refine sentence-level representations has positive effects also on the quality of single token embeddings. Figure \ref{fig:wic2} shows a comparison between Sheared-Llama and Llama 2 in terms of the impact of bidirectional attention, and it can be seen that the larger model is following a similar pattern, although the accuracy scores have a less sharp decrease. Similarly to Sheared-Llama, the application of contrastive learning greatly helps model performance (see Appendix \ref{sec:appendix_result} for more detailed results).



\begin{figure}[ht]
    \centering
    \includegraphics[width=0.45\textwidth]{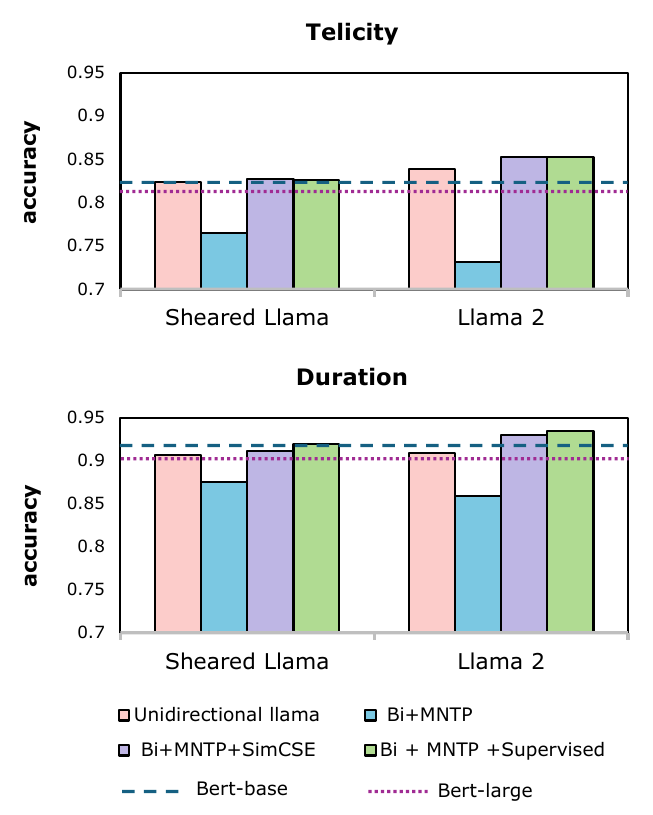} 
    \caption{Results of verb telicity and duration (Task 2).}
    \label{fig:telicity}
\end{figure}

In Figure \ref{fig:telicity} we can see the scores for the \textit{\textbf{lexical aspect classification}}, where we can observe that, for telicity, the basic versions of Llama are already performing well and on par with the bidirectional baselines, while they fall short of BERT Base in modeling verb duration. As we observed before, bidirectional training alone with Bi + MNTP actually makes the models less accurate, whereas contrastive learning techniques consistently improve their performance. At a glance, it can be noticed that Llama 2 with either contrastive training type is better than both BERT baselines in both subtasks.


\textit{\textbf{Finding 2}: Autoregressive models perform similarly to bidirectional ones on regression probing tasks for norms prediction.} 

\begin{figure*}[ht]
    \centering
    \includegraphics[width=0.82\textwidth]{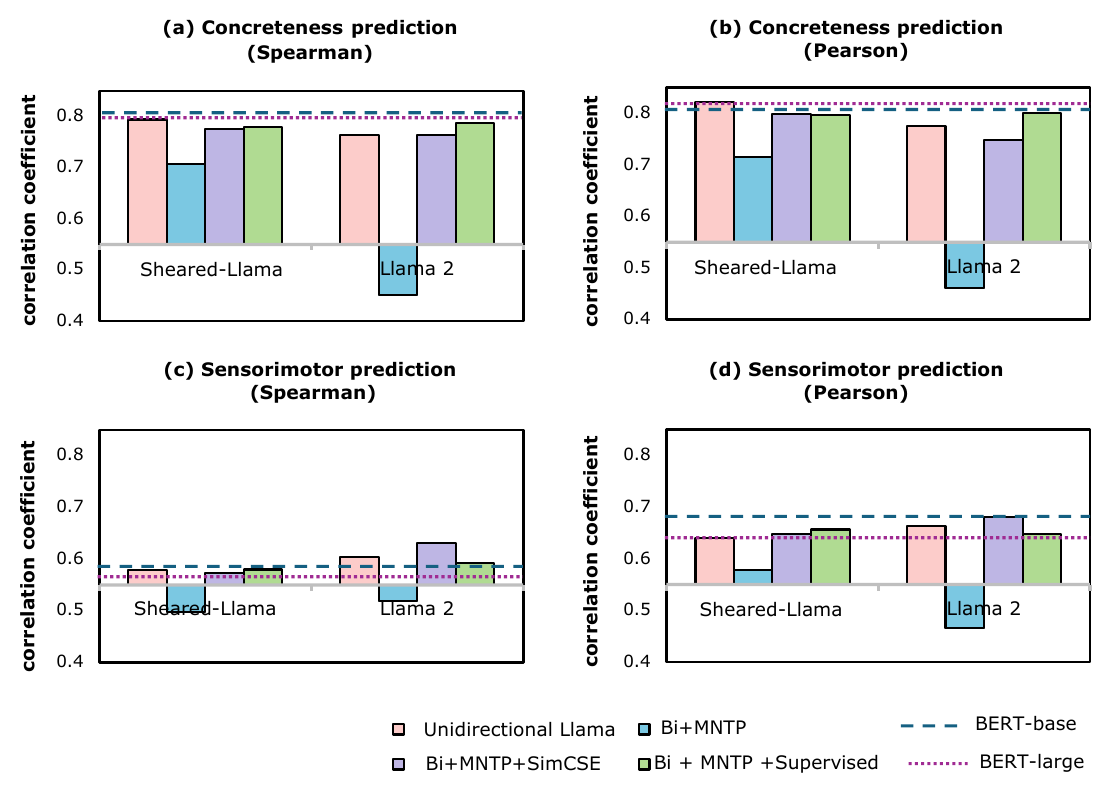} 
    \caption{(a)-(b) show results for predicting noun concreteness (Task 3), while (c)-(d) display average results for predicting 11 sensorimotor dimensions of words (Task 4). Metrics include Spearman and Pearson correlation coefficients.}
    \label{fig:concrete_sensor}
\end{figure*}

Moving to the regression tasks, e.g. concreteness prediction on \textit{\textbf{CONcreTEXT}} and modeling of \textbf{\textit{contextualized sensorimotor norms}} (Figure~\ref{fig:concrete_sensor}), we can immediately see that base Llama models already achieve high correlations with human mean ratings, with no significant disadvantage to the bidirectional competitors. This contradicts previous findings claiming that autoregressive LLMs are not optimal when it comes to modeling word semantics \citep{qorib2024decoder}. 

It should not be underestimated, however, the specificity of the task at hand: while most probing tasks presented in the literature focus on discrete distinctions (e.g. grammatical vs. ungrammatical, semantically plausible vs. implausible), there is only limited comparative evidence about LLM performance when it comes to modeling fine-grained, continuous judgements of semantic properties. Interestingly, recent studies reported that LLMs are able to faithfully reproduce human ratings of concreteness and sensory norms via prompting GPT-4 \citep{xu2023does,martinez2025using}, and thus it is possible that such semantic features are robustly encoded also in other autoregressive architectures (e.g. Llama models).

Once again, Bi + MNTP alone deteriorates embedding quality, and contrastive learning strategies mitigate its negative effects. However, in the concreteness task this does not lead to any consistent improvement over the base models; in the sensorimotor norms task, better correlations can be observed only for the Llama 2 model, and only with the unsupervised contrastive learning strategy.





In addition, it could be noticed about these tasks that in none of the model families size seems to matter too much: Sheared-Llama and BERT Base are often on par or better than the corresponding larger models.

\begin{figure}[!ht]
    \centering
\includegraphics[width=0.45\textwidth]{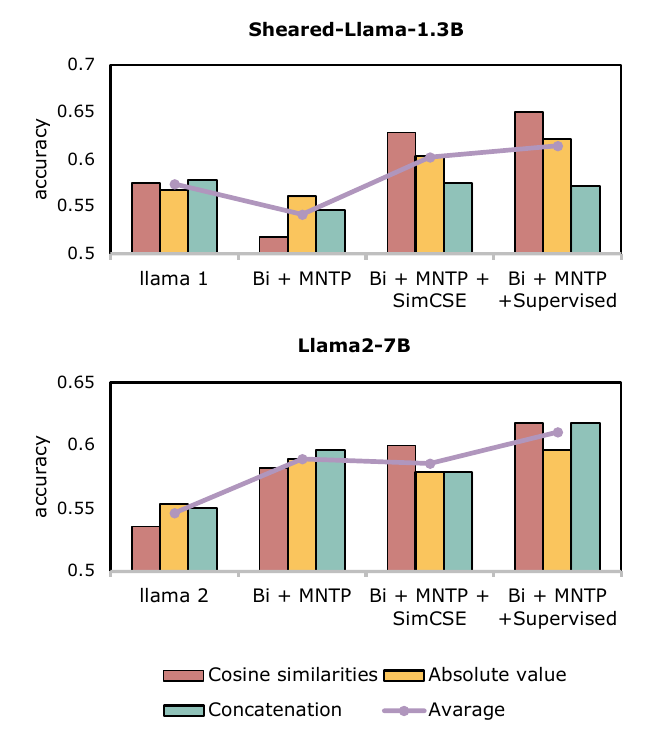} 
    \caption{Results of word sense disambiguation task (Task 5). This task employs three methods of processing two target words’ representation from pair sentences: cosine similarity, absolute difference, and concatenation. The metric is accuracy.}
    \label{fig:wic}
\end{figure}

\noindent \textit{\textbf{Finding 3}: In the sense disambiguation task, contrastive learning methods improve the quality of embeddings from autoregressive models irrespective of the strategy for extracting probe features.}

Given that \textbf{\textit{Word-in-Context}} requires, compared to the other datasets, the combination of two contextualized representations, we tested three different strategies to combine the vectors (Figure~\ref{fig:wic}). No big differences can be seen between those strategies with the base models, and in this case Bi + MNTP has a mixed effect, leading to the usual performance deterioration with Sheared-Llama and to better performance with Llama 2. Interestingly, contrastive learning leads again to best scores in the task, and it can be noticed that after its application the cosine similarities strategy becomes the most consistent one (cosine strategy with Sheared-Llama and either one of contrastive training types are the only combinations outperforming both the BERT baselines). As this strategy is simply based on using vector proximity as the only feature for the probe, the result suggests that refining representations for sentence-level tasks with contrastive learning also leads to more meaningful distances between token-level embeddings in the vector space.


\begin{figure*}[t]
  \includegraphics[width=0.48\linewidth]{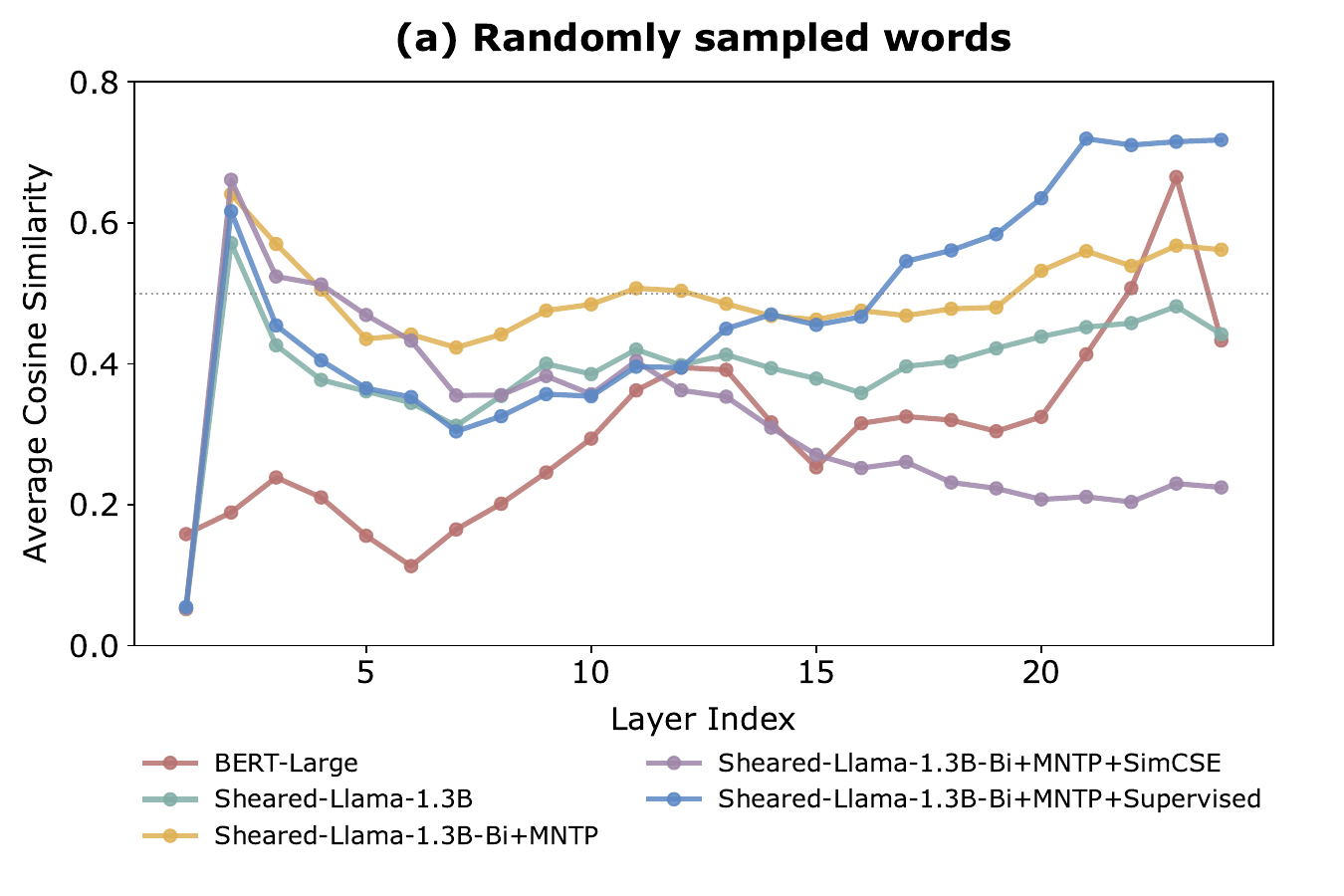} \hfill
  \includegraphics[width=0.48\linewidth]{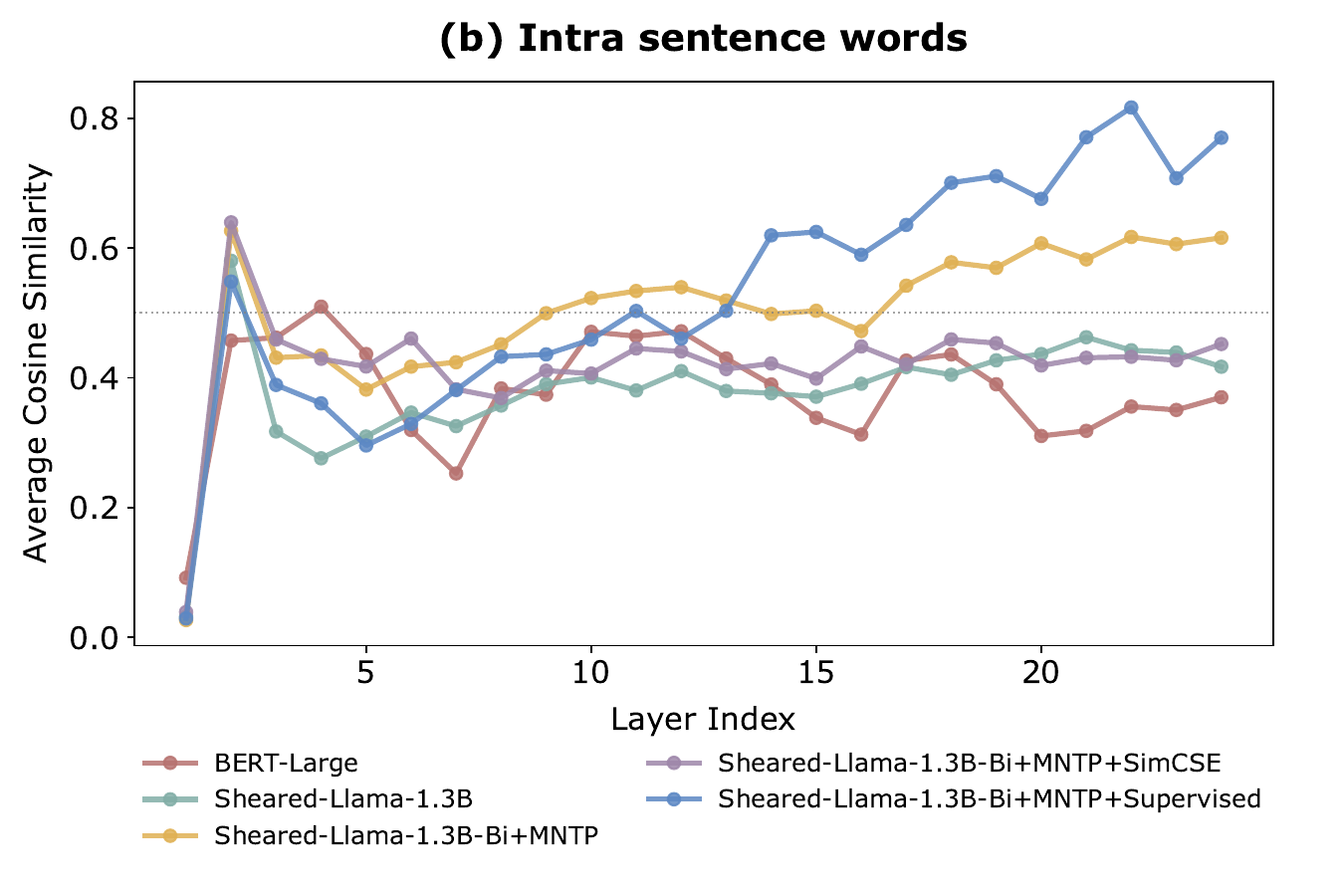}
  \caption {(a) represents the experimental results with randomly sampled words to examine the anisotropy; (b) illustrates the variation in anisotropy of words within the same sentence as the layer depth of the LLMs increases.}
  \label{fig:aniso}
\end{figure*}

\subsection{Anisotropy in the Embedding Space}
Anisotropy, a known issue in pre-trained language models, has been attributed to the disproportionate influence of rare tokens in the negative direction of hidden states within likelihood-maximizing models~\citep{Wang2020Improving,gao2019representationdegenerationproblemtraining}. However, in the literature, bidirectional attention models were shown to exhibit lower anisotropy compared to autoregressive models~\citep{ethayarajh2019contextual}, possibly due to the impact of the attention mechanisms. Based on this, we speculate that bidirectional attention may alleviate the issue of anisotropy in the embeddings of autoregressive models.

Therefore, we extract the embeddings for the target words in the WIC dataset ~\citep{pilehvar-camacho-collados-2019-wic}, randomly sample the representation of all target words 1000 times, and then calculate the average cosine similarity. The result is presented in Figure~\ref{fig:aniso} a): neither BERT nor Llama produces isotropic word representations.  Additionally, all Llama models have a sudden sharp increase in the cosine similarities in the second layer, followed by a decrease. 

In terms of broad trends from shallow to deep layers, Llama models are relatively stable in the central layers and then the similarities increase again towards the last layers, with the exception for unsupervised contrastive learning Llama. The observed impact of the bidirectional attention mechanism contradicts our hypothesis: Sheared-Llama-1.3B with bidirectional attention exhibits higher global similarity across all hidden states compared to its unidirectional counterpart. While the model combining bidirectional attention and supervised contrastive learning is the one with the highest degree of anisotropy, the version with unsupervised contrastive learning is the most successful in reducing average similarities, showing an overall decreasing trend in the later layers.
As for BERT, the similarities show a constant upward trend while moving from the earlier to the later layers, which align with the previous analysis, and showing a spike in the degree of anisotropy in the very last ones \citep{ethayarajh2019contextual}.



On the basis of such findings, we would recommend that for tasks involving an unsupervised evaluation based on vector-space similarity, researchers adopt LLM with bidirectional training and unsupervised contrastive learning, since it seems to be the most robust combination against the anisotropy issue.
On the other hand, our results also show that highly anisotropic representations do not necessarily have a negative impact on supervised tasks. In Figure~\ref{fig:aniso} a) it can be clearly seen that, whereas unsupervised contrastive learning reduces anisotropy, supervised contrastive learning increases it. However, our experiments on the probing datasets showed that Llama models trained with supervised contrastive learning generally outperform all the other models.

To investigate how word representations within the same sentence evolve from shallow to deep layers of the model, we also extract words from individual sentences and perform layer-wise cosine similarity calculations to quantify intra-sentence anisotropy. 
Figure~\ref{fig:aniso} b) shows that supervised contrastive learning bidirectional and bidirectional-only Llama models exhibit increasing anisotropy across layers, indicating that, in the deeper layers of the model, words from the same sentence tend to converge towards similar representations. This exhibits a trend of increasing anisotropy analogous to what we observed with randomly selected words. In the case of intra-sentence words, however, unsupervised contrastive learning does not significantly decrease the similarities compared to the base model, and the lowest levels of anisotropy are generally achieved by BERT in the late layers.

\section{Conclusion}
This work investigates how bidirectional attention contributes to contextual information encoding in word representations. The main contributions are: 1) providing a fine-grained word-based analysis of LLMs with enabled bidirectional attention, 2) unveiling the trade-off brought by bidirectional attention, and 3) contributing valuable insight into the application of LLM2Vec and the ongoing exploration of using LLMs as text encoders.

Contrary to our expectations, we found that in semantic probing tasks simply enabling bidirectional attention is not sufficient, as it may decrease the model's capability to represent the previous context. However, we observed that across tasks, with the only exception of regression tasks for norms prediction, adding contrastive learning on top of bidirectional attention tends to improve the representation quality of the embeddings extracted from autoregressive LLMs. The fact that the sense disambiguation task of the \textbf{\textit{Word-in-Context}} dataset, after contrastive learning, can be better addressed simply by using the cosine between the vectors suggests that this technique might lead to more meaningful distances between token-level vectors.



Moreover, we further analyzed the anisotropy of embedding representations across various hidden states. We found that bidirectional attention increases representation isotropy across all Llama layers, showing that this feature is not inherently related to undirectional attention. Among the contrastive learning strategies, the unsupervised one seems to improve the anisotropy issue in the vector space, and thus it would be the recommended choice if the goal was to adapt an autoregressive LLM to unsupervised word embedding tasks that are evaluated via vector similarity.



\section*{Limitations}

This probing study reveals that activating bidirectional attention impacts the semantic encoding of word representations, though the underlying mechanisms remain unclear. Additionally, the correspondence between high-dimensional dense vectors in computational models and semantic information warrants further investigation to bridge interpretability gaps in neural representation studies.

Another limitation is that our study is limited to the English language with the Sheared-Llama-1.3B and Llama2-7B models, and bigger models were not tested due to limitation of our computational resources. Different languages and models may yield varying effects on the performance our lexical semantic tasks. We anticipate that future work can expand to diverse languages and models to validate and refine our findings. Likewise, Our approach to extracting contextual word representations, while designed for broad model compatibility, may not be necessarily the optimal one for each model. 

Lastly, while bidirectional attention mechanisms have been shown to enhance performance in text embedding tasks in LLMs, their inherent capabilities were not thoroughly evaluated in our study. However, since our paper focuses on the problem of adapting autoregressive LLMs to embedding-based tasks, we did not evaluate text generation capabilities. To our knowledge, bidirectional attention and contrastive learning are not commonly adopted for text-generation tasks. A previous study~\cite{khosla2025magnetaugmentinggenerativedecoders} demonstrated that enabling bidirectional attention in LLMs may significantly exacerbate text repetition. This occurs because such mechanisms disrupt the LLM's native autoregressive generation, causing it to resemble BERT-like models, which are known to struggle with coherent text production. Future work may include more in-depth explorations of this problem.

\section*{Acknowledgements}
We sincerely thank~\citet{behnamghader2024llm2vec} for their innovative work on LLM2Vec, which provided foundational inspiration for our research. Their open-source contribution of the models on HuggingFace have been invaluable to our work. We also thank anonymous reviewers for their constructive feedback and suggestions, which have significantly helped us to refine this study.

\section*{Ethics Statement}
This study investigates the impact of bidirectional attention mechanisms on semantic feature extraction in LLMs. While the technical contributions aim to advance model interpretability and linguistic capability, we acknowledge broader ethical responsibilities in disseminating and applying these findings. The research is intended to support beneficial applications in natural language understanding, such as improving machine translation and text summarization, where enhanced semantic analysis could yield significant societal value. 




\bibliography{custom}

\appendix

\section{Experimental Setup}
\label{sec:appendix}

\subsection{Dataset Details}
\label{sec:appendix_data}

Table~\ref{tab:dataset_details} details datasets used in experiments.  The hyperlinks are attached to dataset names, which direct to the download page of each dataset.


\subsection{Model Details}
\label{sec:appendix_model}

Table~\ref{tab:model_details} shows used model details in the experiments. Hyperlinks to each model's Hugging Face page are added. 

\subsection{Other Setup Details}
\label{sec:appendix_parameters}

The hyperparameters of the Multilayer Perceptron (MLP) classifier used in our experiments are presented in Table~\ref{tab:hyperpatameters}. All experiments were implemented in Python 3.10 utilizing PyTorch 2.6.0 (CUDA 12.4), Transformers 4.43.1, and PEFT 0.10.0 libraries. To ensure reproducibility across experiments, we set the random seed to a fixed value of 42. To reduce memory usage and computational cost, we employed half-precision (FP16) arithmetic for embedding extraction.

\section{Experimental Results}
\label{sec:appendix_result}
\subsection{Multilayer Perceptron}
\label{sec:appendix_result_mlp}

\Cref{tab:task1_data,tab:classification_data,tab:regressiondata_t3,tab:regressiondata_t4} shows the experimental results of five tasks using multilayer perceptron as the probe model. 

\subsection{Logistic Regression/Linear Regression}
\label{sec:appendix_result_lr}

\Cref{tab:task1_data_lr,tab:classificationdata_lr,tab:regressiondata_lr_t3,tab:regressiondata_lr_t4} shows the experimental results of five tasks using logistic regression or linear regression as the probe model.

\section{Effectiveness of Probing Method}
\label{sec:effective_probing}

As ~\citet{voita-titov-2020-information} pointed out, the probe accuracy can be similar when probing for genuine linguistic labels and probing for random synthetic tasks. To address this concern, we introduced control tasks to evaluate the selectivity of the probes. In control tasks, each label is not genuine but randomly sampled. As ~\citet{hewitt2019designing} suggested, a good probe should achieve higher accuracy on linguistic tasks and lower accuracy on control tasks. We test BERT-base, Sheared-Llama, and Sheared-Llama (Bi + BNTP) on control tasks (MLP classifier). 

Results are shown in~\Cref{tab:controltask1,tab:controltask2,tab:controltask3}: Across all five tasks we examined, the control task consistently showed significantly lower accuracy than the original task, demonstrating that the probes exhibited high selectivity in all five cases.


\section{Task Sensitivity to Context}
\label{sec:non_context}

To assess the importance of the context in the first four tasks (i.e., whether  performance benefits from contextual embeddings), we conducted additional experiments using non-contextual inputs (probed words without surrounding context) and MLP
classifier. As demonstrated in~\Cref{tab:task1withoutcontext,tab:task2withoutcontext}, the results confirm that all four tasks indeed show improved performance with contextual information.

In Task 1 (Table ~\ref{tab:task1withoutcontext}): for each experiment, we input only a single word (the probed word) to the model, rather than an entire sentence. As expected, we found that accuracy is high only when probing the target word directly, with the other words yielding random-guess performance. Notice that in Task 1 our focus is not on target word probing accuracy itself, but rather on how much semantic information about the target word is contained in the embeddings of the other words. Therefore, task 1 can be solved only by contextualized embeddings for an input token other than the probed word.

As for Task 2, Task 3, and Task 4, we also added no-context results (Table~\ref{tab:task2withoutcontext}). We found that results without context are significantly lower than those with context, indicating that contextual information is crucial for prediction in all three tasks.



\begin{table*}
\centering
\begin{tabular}{|l|l|ll|ll|}
\hline
Dataset & Predictive Method & \multicolumn{2}{l|}{Train Data} & \multicolumn{2}{l|}{Test Data}\\ \hline
\begin{tabular}[c]{@{}l@{}}\href{https://github.com/jklafka/context-probes}{Context-probe}\end{tabular} & Classification & \multicolumn{2}{l|}{\begin{tabular}[c]{@{}l@{}} Subject:\\ 4000 sentences\\ Object:\\ 4000 sentences\\ Verb:\\ 8000 sentences\end{tabular}} & \multicolumn{2}{l|}{\begin{tabular}[c]{@{}l@{}} Subject:\\ 1000 sentences\\ Object:\\ 1000 sentences\\ Verb:\\ 2000 sentences\end{tabular}} \\ \hline
\begin{tabular}[c]{@{}l@{}}\href{https://github.com/lenakmeth/telicity_classification}{\makecell[tl]{Lexical aspect binary\\ Classification (telicity/duration)}}\end{tabular} & Classification & \multicolumn{2}{l|}{\begin{tabular}[c]{@{}l@{}} Telicity:\\ 3920 sentences\\ Duration:\\ 2591 sentences\end{tabular}}& \multicolumn{2}{l|}{\begin{tabular}[c]{@{}l@{}} Telicity:\\ 980 sentences\\ Duration:\\ 648 sentences\end{tabular}} \\ \hline
\begin{tabular}[c]{@{}l@{}}\href{https://osf.io/j4dz3/}{CONcreTEXT}\end{tabular} & Regression & \multicolumn{2}{l|}{\begin{tabular}[c]{@{}l@{}} 347 sentences\end{tabular}}  & \multicolumn{2}{l|}{\begin{tabular}[c]{@{}l@{}} 87 sentences\end{tabular}} \\ \hline
\begin{tabular}[c]{@{}l@{}}\href{https://github.com/seantrott/cs_norms}{\makecell[tl]{Contextualized Sensorimotor\\ Norms}}\end{tabular} & Regression & \multicolumn{2}{l|}{\begin{tabular}[c]{@{}l@{}}358 sentences\end{tabular}} & \multicolumn{2}{l|}{\begin{tabular}[c]{@{}l@{}}90 sentences\end{tabular}} \\ \hline
\begin{tabular}[c]{@{}l@{}}\href{https://pilehvar.github.io/wic/}{WiC Dataset}\end{tabular} & Classification & \multicolumn{2}{l|}{\begin{tabular}[c]{@{}l@{}} 1120 sentence pairs\end{tabular}} & \multicolumn{2}{l|}{\begin{tabular}[c]{@{}l@{}} 280 sentence pairs\end{tabular}} \\ \hline
\end{tabular}
\caption{Dataset details and train-test data separation of classifier or regressor training in our experiments. The predictive method applied and data separation between the train and test dataset are demonstrated. Links leading to the webpage of Github or the related recourse are added to the dataset name.}
\label{tab:dataset_details}
\end{table*}

\begin{table*}
\begin{tabular}{ccccccccc}
\toprule
 & T1 & T2-T & T2-D & T3 & T4 & T5-A & T5-S & T5-C \\
\midrule
Structure & {[}20, 10{]} & {[}20, 10{]} & {[}20, 10{]} & {[}40, 20{]} & {[}40, 20{]} & {[}20,10{]} & {[}5,2{]} & {[}20, 10{]} \\
Batch size & 16 & 16 & 16 & 8 & 16 & 16 & 8 & 16 \\
Learning rate & 2e-5 & 2e-5 & 2e-5 & 2e-3 & 2e-3 & 1e-5 & 1e-3 & 1.5e-5\\
\bottomrule
\end{tabular}
\caption{Hyperparameters in Multilayer Perceptron classifier in experiments. All language models utilize the same hyperparameters setting in the same task. The first row indicates Task 1, Task 2 telicity, Task 3 duration, Task 3, Task 4, Task 5 (absolute difference), Task 5 (cosine similarity), and Task 5 (concatenation).}
\label{tab:hyperpatameters}
\end{table*}

\begin{table*}
\normalsize 
\begin{tabular}{|l|l|l|l|}
\hline
Family & Model Link & Parameters & Attention Mechanism \\ \hline
\multicolumn{1}{|c|}{\multirow{2}{*}{BERT}} & \href{https://huggingface.co/google-bert/bert-base-uncased}{bert-base-uncased} & 109M & \multirow{2}{*}{Bidirectional} \\ \cline{2-3}
\multicolumn{1}{|c|}{} & \href{https://huggingface.co/google-bert/bert-large-uncased}{bert-large-uncased} & 335M &  \\ \hline
\multirow{4}{*}{Llama 1} & \href{https://huggingface.co/princeton-nlp/Sheared-LLaMA-1.3B}{Sheared-LLaMA-1.3B} & \multirow{4}{*}{1.3B} & Unidirectional \\ \cline{2-2} \cline{4-4} 
 & \href{https://huggingface.co/McGill-NLP/LLM2Vec-Sheared-LLaMA-mntp}{LLM2Vec-Sheared-LLaMA-mntp} &  & \multirow{3}{*}{Bidirectional} \\ \cline{2-2}
 & \href{https://huggingface.co/McGill-NLP/LLM2Vec-Sheared-LLaMA-mntp-unsup-simcse}{LLM2Vec-Sheared-LLaMA-mntp-unsup-simcse} &  &  \\ \cline{2-2}
 & \href{https://huggingface.co/McGill-NLP/LLM2Vec-Sheared-LLaMA-mntp-supervised}{LLM2Vec-Sheared-LLaMA-mntp-supervised} &  &  \\ \hline
\multirow{4}{*}{Llama 2} & \href{https://huggingface.co/meta-llama/Llama-2-7b-chat-hf}{Llama-2-7b-hf} & \multirow{4}{*}{7B} & Unidirectional \\ \cline{2-2} \cline{4-4} 
 & \href{https://huggingface.co/McGill-NLP/LLM2Vec-Llama-2-7b-chat-hf-mntp}{LLM2Vec-Llama-2-7b-chat-hf-mntp} &  & \multirow{3}{*}{Bidirectional} \\ \cline{2-2}
 & \href{https://huggingface.co/McGill-NLP/LLM2Vec-Llama-2-7b-chat-hf-mntp-unsup-simcse}{LLM2Vec-Llama-2-7b-chat-hf-mntp-unsup-simcse} &  &  \\ \cline{2-2}
 & \href{https://huggingface.co/McGill-NLP/LLM2Vec-Llama-2-7b-chat-hf-mntp-supervised}{LLM2Vec-Llama-2-7b-chat-hf-mntp-supervised} &  &  \\ \hline
\end{tabular}

\caption{Model details of used models, parameters details, and enabled attention mechanism in our experiments. Links leading to the Hugging Face page of each model are added to the hyperlinks.}
\label{tab:model_details}
  
\end{table*}

\begin{table*}
\begin{tabular}{ccccccc}
\toprule
Model & Subtask & Index 1 & Index 2 & Index 3 & Index 4 & Index 5 \\
\midrule
\multirow{4}{*}{BERT-base} & subject animacy & 0.942 & 1.000 & 0.966 & 0.944 & 0.944 \\
 & verb causative & 0.989 & 0.976 & 0.999 & 0.990 & 0.989 \\
 & verb dynamic & 0.998 & 0.979 & 1.000 & 0.999 & 0.984 \\
 & object animacy & 0.942 & 0.870 & 0.914 & 0.948 & 0.982 \\
\midrule
\multirow{4}{*}{BERT-large} & subject animacy & 0.967 & 0.998 & 0.965 & 0.969 & 0.939 \\
 & verb causative & 0.989 & 0.945 & 0.998 & 0.990 & 0.968 \\
 & verb dynamic & 0.992 & 0.975 & 1.000 & 0.991 & 0.971 \\
 & object animacy & 0.932 & 0.847 & 0.889 & 0.940 & 0.987 \\
 \midrule
\multirow{4}{*}{Sheared-Llama-1.3B} & subject animacy & 0.547$\dagger$ & 1.000 & 0.994 & 0.996 & 0.943 \\
 & verb causative & 0.611$\dagger$ & 0.534$\dagger$ & 1.000 & 0.995 & 0.987 \\
 & verb dynamic & 0.515$\dagger$ & 0.494$\dagger$ & 1.000 & 0.998 & 0.973 \\
 & object animacy & 0.539$\dagger$ & 0.501$\dagger$ & 0.508$\dagger$ & 0.507$\dagger$ & 1.000 \\
 \midrule
\multirow{4}{*}{\begin{tabular}[c]{@{}c@{}}Sheared-Llama-1.3B\\  (Bi + MNTP)\end{tabular}} & subject animacy & 0.976 & 0.976$\star$ & 0.916$\star$ & 0.826$\star$ & 0.766$\star$ \\
 & verb causative & 0.958 & 0.968 & 0.991$\star$ & 0.960$\star$ & 0.939$\star$ \\
 & verb dynamic & 0.972 & 0.974 & 0.983$\star$ & 0.954$\star$ & 0.935$\star$ \\
 & object animacy & 0.850 & 0.877 & 0.960 & 0.995 & 0.999$\star$ \\
 \midrule
\multirow{4}{*}{\begin{tabular}[c]{@{}c@{}}Sheared-Llama-1.3B\\ (Bi + MNTP + SimCSE)\end{tabular}} & subject animacy & 0.995 & 1.000 & 0.949 & 0.949 & 0.957 \\
 & verb causative & 0.993 & 0.999 & 1.000 & 0.995 & 0.997 \\
 & verb dynamic & 0.999 & 1.000 & 1.000 & 0.997 & 0.997 \\
 & object animacy & 0.944 & 0.956 & 0.978 & 0.996 & 1.000 \\
 \midrule
\multirow{4}{*}{\begin{tabular}[c]{@{}c@{}}Sheared-Llama-1.3B\\ (Bi + MNTP +Supervised)\end{tabular}} & subject animacy & 0.988 & 1.000 & 0.980 & 0.983 & 0.993 \\
 & verb causative & 0.998 & 0.995 & 1.000 & 0.999 & 0.999 \\
 & verb dynamic & 1.000 & 0.999 & 1.000 & 0.999 & 0.997 \\
 & object animacy & 0.993 & 0.995 & 0.989 & 0.999 & 1.000 \\
 \midrule

\multirow{4}{*}{Llama2-7B} & subject animacy & 0.547$\dagger$ & 1.000 & 0.995 & 0.996 & 0.962 \\
 & verb causative & 0.611$\dagger$ & 0.526$\dagger$ & 1.000 & 1.000 & 0.997 \\
 & verb dynamic & 0.515$\dagger$ & 0.515$\dagger$ & 1.000 & 1.000 & 0.997 \\
 & object animacy & 0.544$\dagger$ & 0.496$\dagger$ & 0.497$\dagger$ & 0.510$\dagger$ & 1.000 \\
 \midrule
 
\multirow{4}{*}{\begin{tabular}[c]{@{}c@{}}Llama2-7B\\ (Bi + MNTP)\end{tabular}} & subject animacy & 0.999 & 0.998$\star$ & 0.978$\star$ & 0.908$\star$ & 0.897$\star$ \\
 & verb causative & 0.979 & 0.990 & 1.000 & 0.986$\star$ & 0.978$\star$ \\
 & verb dynamic & 0.985 & 0.990 & 0.996$\star$ & 0.983$\star$ & 0.978$\star$ \\
 & object animacy & 0.901 & 0.941 & 0.994 & 0.999 & 0.997$\star$ \\
 \midrule

 \multirow{4}{*}{\begin{tabular}[c]{@{}c@{}}Llama2-7B\\ (Bi + MNTP + SimCSE)\end{tabular}} & subject animacy & 0.993 & 1.000 & 0.984 & 0.982 & 0.994 \\
 & verb causative & 0.998 & 1.000 & 1.000 & 0.999 & 0.999 \\
 & verb dynamic & 0.998 & 1.000 & 1.000 & 0.999 & 1.000 \\
 & object animacy & 0.991 & 0.996 & 0.988 & 0.996 & 1.000 \\
 \midrule

 \multirow{4}{*}{\begin{tabular}[c]{@{}c@{}}Llama2-7B\\ (Bi + MNTP + Supervised)\end{tabular}} & subject animacy & 1.000 & 1.000 & 0.991 & 0.987 & 0.999 \\
 & verb causative & 1.000 & 0.999 & 1.000 & 0.999 & 0.999 \\
 & verb dynamic & 1.000 & 1.000 & 1.000 & 1.000 & 0.997 \\
 & object animacy & 0.983 & 0.990 & 0.986 & 0.999 & 1.000 \\
\bottomrule
\end{tabular}
\caption{Results of Task 1 (context probes, Multilayer Perceptron). Numbers with $\dagger$ give evidence that LLMs can not access the subsequent context, and numbers with $\star$ show that bidirectional attention weakens preceding context utilization.}
\label{tab:task1_data}
\end{table*}

\clearpage

\begin{table*}
\centering
\begin{tabular}{lllllll}
\toprule
\textbf{Model} & \textbf{T2-T} & \textbf{T2-D} & \textbf{T5-A} & \textbf{T5-S} & \textbf{T5-C} & \textbf{T5-Avg}\\
\midrule
BERT-base & 0.8286 & 0.9167 & 0.6071 & 0.6179 & 0.5857 & 0.6036\\
BERT-large & 0.8235 & 0.9059 & \textbf{0.6429} & \textbf{0.6786} & \underline{0.5393} & \textbf{0.6203}\\
Sheared-Llama-1.3B & 0.8235 & 0.9074 & 0.5679 & 0.5750 & 0.5786 & 0.5738\\\addlinespace
Sheared-Llama-1.3B-Bi+MNTP & 0.7653 & 0.8750 & 0.5607 & \underline{0.5179} & 0.5464 & \underline{0.5417}\\
Sheared-Llama-1.3B-Bi+MNTP+SimCSE & 0.8276 & 0.9120 & 0.6036 & 0.6286 & 0.5750 & 0.6024\\
Sheared-Llama-1.3B-Bi+MNTP+Supv. & 0.8265 & 0.9198 & 0.6214 & 0.6500 & 0.5714 & 0.6143\\\addlinespace
Llama2-7B & 0.8388 & 0.9090 & \underline{0.5536} & 0.5357 & 0.5500 & 0.5464\\
Llama2-7B-Bi+MNTP & \underline{0.7316} & \underline{0.8596} & 0.5893 & 0.5821 & 0.5964 & 0.5893\\
Llama2-7B-Bi+MNTP+SimCSE & \textbf{0.8531} & 0.9306 & 0.5786 & 0.6000 & 0.5786 & 0.5857\\
Llama2-7B-Bi+MNTP+Supv. & \textbf{0.8531} & \textbf{0.9352} & 0.5964 & 0.6179 & \textbf{0.6179}  & 0.6107\\
\bottomrule
\end{tabular}
\caption{Classification accuracy of Task 2 and Task 5 (Multilayer Perceptron): T2-T (Task 2 telicity subtask), T2-D (Task 2 duration subtask), T5-A/S/C (Task 5 with the absolute difference/cosine similarity/concatenation.) “Supv.” is the abbreviation of “Supervised”. The bolded numbers indicate the highest performances among models in the same column, while numbers with underlines mean the weakest performance.}
\label{tab:classification_data}
\end{table*}

\begin{table*}
\centering
\begin{tabular}{lllll}
\toprule
\textbf{Model} & \textbf{T3-MSE}  & \textbf{T3-R\textsuperscript{2}} & \textbf{T3-Pearson} & \textbf{T3-Spearman}\\ \midrule
BERT-base & 0.6758 & 0.6612 & 0.8215 & 0.7971\\
BERT-large & \textbf{0.6639} & \textbf{0.6671} & \textbf{0.8219} & \textbf{0.7946}\\\addlinespace
Sheared-Llama-1.3B & 0.7336 & 0.6322 & \textbf{0.8219} & \textbf{0.7946}\\ 
Sheared-Llama-1.3B-Bi+MNTP & 0.9808 & 0.5082 & 0.7157 & 0.7079\\ 
Sheared-Llama-1.3B-Bi+MNTP+SimCSE & 0.7383 & 0.6298 & 0.7983 & 0.7768\\ 
 Sheared-Llama-1.3B-Bi+MNTP+Supv. &  0.6926 & 0.6527 & 0.7958 & 0.7795 \\\addlinespace 
Llama2-7B & 0.8213 & 0.5882 & 0.7751 & 0.7635 \\
Llama2-7B-Bi+MNTP & \underline{1.5857} & \underline{0.2049} & \underline{0.4616} & \underline{0.4503} \\
Llama2-7B-Bi+MNTP+SimCSE & 0.8872 & 0.5552 & 0.7477 & 0.7628 \\
 Llama2-7B-Bi+MNTP+Supv. & 0.7241 & 0.6370 & 0.8012 & 0.7879 \\ \bottomrule
\end{tabular}
\caption{Regression results of Task 3 (concreteness prediction, Multilayer Perceptron).  “Supv.” is the abbreviation of “Supervised”.}
\label{tab:regressiondata_t3}
\end{table*}

\begin{table*}
\centering
\begin{tabular}{lllll}
\toprule
\textbf{Model} & \textbf{T4-MSE} & \textbf{T4-R\textsuperscript{2}} & \textbf{T4-Pearson} & \textbf{T4-Spearman}\\ \midrule
BERT-base & 0.3738 & 0.4065 & 0.6706 & 0.5916 \\
BERT-large & 0.3905 & 0.3483 & 0.6300 & 0.5567\\\addlinespace
Sheared-Llama-1.3B & 0.3884 & 0.3664 & 0.6408 & 0.5785\\
Sheared-Llama-1.3B-Bi+MNTP & 0.4626 & \underline{-0.4610} & 0.5788 & \underline{0.4976}\\
Sheared-Llama-1.3B-Bi+MNTP+SimCSE & 0.3766 & 0.4215 & 0.6484 & 0.5730\\ 
Sheared-Llama-1.3B-Bi+MNTP+Supv. & 0.3835 & 0.3798 & 0.6572 & 0.5798 \\ \addlinespace
Llama2-7B & 0.3854 & 0.4154 & 0.6626 & 0.6040 \\
Llama2-7B-Bi+MNTP & \underline{0.5505} & 0.1782 & \underline{0.4659} & 0.5179 \\
Llama2-7B-Bi+MNTP+SimCSE & \textbf{0.3616} & \textbf{0.4240} & \textbf{0.6806} & \textbf{0.6317} \\
 Llama2-7B-Bi+MNTP+Supv.& 0.3903 & 0.3841 & 0.6486 & 0.5927 \\ \bottomrule
\end{tabular}
\caption{Average regression results of Task 4 (sensorimotor prediction, Multilayer Perceptron)  “Supv.” is the abbreviation of “Supervised”.}
\label{tab:regressiondata_t4}
\end{table*}

\begin{table*}
\begin{tabular}{ccccccc}
\toprule
Model & Subtask & Index 1 & Index 2 & Index 3 & Index 4 & Index 5 \\
\midrule
\multirow{4}{*}{BERT-base} & subject animacy & 0.925 & 0.999 & 0.958 & 0.925 & 0.936 \\
 & verb causative & 0.975 & 0.959 & 0.994 & 0.975 & 0.980 \\
 & verb dynamic & 0.989 & 0.984 & 1.000 & 0.989 & 0.982 \\
 & object animacy & 0.931 & 0.835 & 0.913 & 0.931 & 0.984 \\
 \midrule
\multirow{4}{*}{BERT-large} & subject animacy & 0.944 & 0.997 & 0.961 & 0.944 & 0.929 \\
 & verb causative & 0.988 & 0.931 & 0.998 & 0.988 & 0.970 \\
 & verb dynamic & 0.988 & 0.974 & 0.998 & 0.988 & 0.969 \\
 & object animacy & 0.925 & 0.810 & 0.871 & 0.925 & 0.988 \\
  \midrule
\multirow{4}{*}{Sheared-Llama-1.3B} & subject animacy & 0.547 $\dagger$ & 1.000 & 0.999 & 0.998 & 0.959 \\
 & verb causative & 0.611 $\dagger$ & 0.527 $\dagger$ & 1.000 & 0.997 & 0.996 \\
 & verb dynamic & 0.515 $\dagger$ & 0.500 $\dagger$ & 1.000 & 1.000 & 0.991 \\
 & object animacy & 0.539 $\dagger$ & 0.492 $\dagger$ & 0.534 $\dagger$ & 0.479 $\dagger$ & 1.000 \\
  \midrule
\multirow{4}{*}{\begin{tabular}[c]{@{}c@{}}Sheared-Llama-1.3B\\  (Bi + MNTP)\end{tabular}} & subject animacy & 0.979 & 0.983 $\star$ & 0.912 $\star$ & 0.823 $\star$ & 0.765 $\star$ \\
 & verb causative & 0.962 & 0.969 & 0.994 $\star$ & 0.966 $\star$ & 0.943 $\star$ \\
 & verb dynamic & 0.971 & 0.972 & 0.987 $\star$ & 0.972 $\star$ & 0.945 $\star$ \\
 & object animacy & 0.882 & 0.887 & 0.956 & 0.988 & 0.995 $\star$ \\
  \midrule
\multirow{4}{*}{\begin{tabular}[c]{@{}c@{}}Sheared-Llama-1.3B\\ (Bi + MNTP + SimCSE)\end{tabular}} & subject animacy & 0.995 & 1.000 & 0.961 & 0.947 & 0.962 \\
 & verb causative & 0.996 & 0.999 & 1.000 & 0.999 & 0.998 \\
 & verb dynamic & 1.000 & 1.000 & 1.000 & 0.998 & 0.999 \\
 & object animacy & 0.952 & 0.970 & 0.986 & 0.996 & 1.000 \\
  \midrule
\multirow{4}{*}{\begin{tabular}[c]{@{}c@{}}Sheared-Llama-1.3B\\ (Bi + MNTP +Supervised)\end{tabular}} & subject animacy & 0.970 & 1.000 & 0.990 & 0.991 & 0.994 \\
 & verb causative & 0.986 & 0.998 & 1.000 & 1.000 & 0.998 \\
 & verb dynamic & 0.996 & 1.000 & 1.000 & 1.000 & 0.999 \\
 & object animacy & 0.994 & 0.997 & 0.992 & 0.999 & 1.000 \\
  \midrule

\multirow{4}{*}{Llama2-7B} & subject animacy & 0.547$\dagger$ & 1.000 & 0.998 & 0.998 & 0.977 \\
 & verb causative & 0.611$\dagger$ & 0.528$\dagger$ & 1.000 & 1.000 & 0.994 \\
 & verb dynamic & 0.512 $\dagger$ & 0.490$\dagger$ & 1.000 & 1.000 & 0.996 \\
 & object animacy & 0.544$\dagger$ & 0.496$\dagger$ & 0.507$\dagger$ & 0.505$\dagger$ & 0.999 \\
 \midrule
 
\multirow{4}{*}{\begin{tabular}[c]{@{}c@{}}Llama2-7B\\ (Bi + MNTP)\end{tabular}} & subject animacy & 0.997 & 0.993$\star$ & 0.986$\star$ & 0.949$\star$ & 0.936$\star$ \\
 & verb causative & 0.979 & 0.990 & 0.997$\star$ & 0.994$\star$ & 0.983$\star$ \\
 & verb dynamic & 0.988 & 0.992 & 0.997$\star$  & 0.990$\star$  & 0.985$\star$  \\
 & object animacy & 0.932 & 0.958 & 0.989 & 0.993 & 0.994$\star$  \\
 \midrule

 \multirow{4}{*}{\begin{tabular}[c]{@{}c@{}}Llama2-7B\\ (Bi + MNTP + SimCSE)\end{tabular}} & subject animacy & 0.998 & 1.000 & 0.993 & 0.984 & 0.993 \\
 & verb causative & 1.000 & 1.000 & 1.000 & 1.000 & 1.000 \\
 & verb dynamic & 1.000 & 1.000 & 1.000 & 1.000 & 1.000 \\
 & object animacy & 0.989 & 0.992 & 0.990 & 0.997 & 1.000 \\
 \midrule

 \multirow{4}{*}{\begin{tabular}[c]{@{}c@{}}Llama2-7B\\ (Bi + MNTP + Supervised)\end{tabular}} & subject animacy & 1.000 & 1.000 & 0.993 & 0.988 & 0.999 \\
 & verb causative & 1.000 & 0.999 & 1.000 & 1.000 & 0.998 \\
 & verb dynamic & 1.000 & 1.000 & 1.000 & 0.999 & 0.999 \\
 & object animacy & 0.982 & 0.991 & 0.989 & 0.999 & 1.000 \\
 
\bottomrule
\end{tabular}
\caption{Results of Task 1 (context probes, Logistic Regression). Numbers with $\dagger$ give evidence that LLMs can not access the subsequent context, and numbers with $\star$ show that bidirectional attention weakens preceding context utilization.}
\label{tab:task1_data_lr}
\end{table*}

\begin{table*}
\centering
\begin{tabular}{lllllll}
\toprule
\textbf{Model} & \textbf{T2-T} & \textbf{T2-D} & \textbf{T5-A} & \textbf{T5-S} & \textbf{T5-C} & \textbf{T5-Avg}\\
\midrule
BERT-base & 0.7765 & 0.8904 & \textbf{0.5929} & 0.6179 & \textbf{0.5607} & 0.5905\\
BERT-large & 0.7837 & 0.8796 & 0.5357 & \textbf{0.6643} & \underline{0.5000} & 0.5666\\\addlinespace
Sheared-Llama-1.3B & 0.7571 & 0.9059 & 0.4964 & 0.5357 & 0.5107 & 0.5142\\
Sheared-Llama-1.3B-Bi+MNTP & 0.7010 & 0.8611 & 0.5321 & 0.5500 & 0.5071 & 0.5297\\
Sheared-Llama-1.3B-Bi+MNTP+SimCSE & 0.7745 & 0.9090 & 0.5500 & 0.6321 & 0.5357 & 0.5726\\
Sheared-Llama-1.3B-Bi+MNTP+Supv. & 0.7918 & 0.9136 & 0.5821 & 0.6429 & 0.5500 & \textbf{0.5917}\\\addlinespace
Llama2-7B & 0.8061 & 0.9151 & \underline{0.4679} & \underline{0.5107} & \underline{0.5000} & \underline{0.4929}\\
Llama2-7B-Bi+MNTP & \underline{0.6704} & \underline{0.8596} & 0.5179 & 0.5500 & 0.5250 & 0.5310\\
Llama2-7B-Bi+MNTP+SimCSE & 0.8194 & \textbf{0.9244} & 0.5571 & 0.6000 & 0.5429 & 0.5667\\
Llama2-7B-Bi+MNTP+Supv. & \textbf{0.8357} & 0.9198 & 0.5500 & 0.6393 & 0.5321 & 0.5738\\
\bottomrule
\end{tabular}
\caption{Classification accuracy of Task 2 and Task 5 (Logistic Regression): T2-T (Task 2 telicity subtask), T2-D (Task 2 duration subtask), T5-A/S/C (Task 5 with the absolute difference/cosine similarity/concatenation.) “Supv.” is the abbreviation of “Supervised”.}
\label{tab:classificationdata_lr}
\end{table*}

\begin{table*}
\centering
\begin{tabular}{lllll}
\toprule
\textbf{Model} & \textbf{T3-MSE} & \textbf{T3-R\textsuperscript{2}} & \textbf{T3-Pearson} & \textbf{T3-Spearman}\\ \midrule
BERT-base & 1.2702 & 0.3631 & 0.6967 & 0.6774\\
BERT-large &  0.9944 & 0.5014 & 0.7457 & 0.7319\\\addlinespace
Sheared-Llama-1.3B & 0.8761 & 0.5607 & 0.7778 & 0.7588 \\
Sheared-Llama-1.3B-Bi+MNTP & 1.1636  & 0.4166 & 0.6866 & 0.6792\\ 
Sheared-Llama-1.3B-Bi+MNTP+SimCSE & 0.7638 & 0.6171 & 0.7914 & 0.7668\\ 
Sheared-Llama-1.3B-Bi+MNTP+Supv. & 0.7269 & 0.6356 & 0.8107 & 0.7905\\ \addlinespace
Llama2-7B & 0.8379 & 0.5799 & 0.7804 & 0.7638 \\ 
Llama2-7B-Bi+MNTP & \underline{1.5123} & \underline{0.2418} & \underline{0.5370} & \underline{0.5435}\\ 
Llama2-7B-Bi+MNTP+SimCSE & 0.8108 & 0.5935 & 0.7895 & 0.7724 \\ 
Llama2-7B-Bi+MNTP+Supv. & \textbf{0.6597} & \textbf{0.6692} & \textbf{0.8375} & \textbf{0.8190}\\ \bottomrule
\end{tabular}
\caption{Regression results of Task 3 (concreteness prediction, Linear Regression).  “Supv.” is the abbreviation of “Supervised”.}
\label{tab:regressiondata_lr_t3}
\end{table*}

\begin{table*}
\centering
\begin{tabular}{lllll}
\toprule
\textbf{Model} & \textbf{T4-MSE} & \textbf{T4-R\textsuperscript{2}} & \textbf{T4-Pearson} & \textbf{T4-Spearman}\\ \midrule
BERT-base & \underline{0.6625} & 0.0215 & 0.5680 & 0.4917 \\ 
BERT-large & 0.5907 & 0.0862 & 0.5695 & 0.5100\\ \addlinespace
Sheared-Llama-1.3B & 0.4152 & 0.3446 & 0.6484 & 0.5859\\ 
Sheared-Llama-1.3B-Bi+MNTP & 0.6371 & 0.0201& 0.5185 & 0.4603\\ 
Sheared-Llama-1.3B-Bi+MNTP+SimCSE &  0.4029 & 0.3829 & 0.6662 & 0.6085\\ 
Sheared-Llama-1.3B-Bi+MNTP+Supv. & 0.4554 & 0.2950 & 0.6304 & 0.5622 \\\addlinespace 
Llama2-7B & 0.3827 & 0.3994 & 0.6636 & 0.5944 \\ 
Llama2-7B-Bi+MNTP & 0.6598 & \underline{0.0121} & \underline{0.4925} & \underline{0.4301} \\
Llama2-7B-Bi+MNTP+SimCSE & 0.3560 & 0.4508 & 0.6952 & \textbf{0.6327} \\
Llama2-7B-Bi+MNTP+Supv. & \textbf{0.3547} & \textbf{0.4577} & \textbf{0.6965} & 0.6271 \\ \bottomrule
\end{tabular}
\caption{Average regression results of Task 4 (sensorimotor prediction, Linear Regression). “Supv.” is the abbreviation of “Supervised”.}
\label{tab:regressiondata_lr_t4}
\end{table*}

\begin{table*}
\centering
\begin{tabular}{lllllll}
\toprule
\textbf{Model} & \textbf{Task} & \textbf{Index1} & \textbf{Index2} & \textbf{Index3} & \textbf{Index4} & \textbf{Index5} \\ 
\midrule
BERT-base & control & 0.518 & 0.494 & 0.529 & 0.519 & 0.520 \\
 & original & 0.942 & 1.000 & 0.966 & 0.944 & 0.944 \\
Sheared-Llama & control & 0.497 & 0.520 & 0.490 & 0.530 & 0.510 \\
 & original & 0.547 & 1.000 & 0.994 & 0.996 & 0.943 \\
Sheared-Llama (Bi+BNTP) & control & 0.510 & 0.527 & 0.485 & 0.503 & 0.493 \\
 & original & 0.976 & 0.976 & 0.916 & 0.826 & 0.766\\ \bottomrule
\end{tabular}
\caption{This table presents a comparison of accuracy between the control task and the original task in Task 1 (subject animacy).}
\label{tab:controltask1}
\end{table*}


\begin{table*}[ht]
\centering
\begin{tabular}{llllll}
\toprule
\textbf{Model} & \textbf{Task} & \textbf{Task3-P} & \textbf{Task3-S} & \textbf{Task4-P} & \textbf{Task4-S} \\
\midrule
BERT-base & control & 0.0649 & 0.0810 & 0.0421 & 0.0417 \\
 & original & 0.8215 & 0.7971 & 0.6706 & 0.5916 \\
Sheared-Llama & control & -0.0798 & -0.0739 & -0.0097 & -0.0091 \\
 & original & 0.8219 & 0.7946 & 0.6408 & 0.5785 \\
Sheared-Llama (Bi + BNTP) & control & 0.1214 & 0.1228 & 0.0437 & 0.0337 \\
 & original & 0.7157 & 0.7079 & 0.5788 & 0.4976\\
\bottomrule
\end{tabular}
\caption{This table presents a comparison of Pearson and Spearman correlation coefficient between the control task and the original task in Task 3 and 4. P and S represent Pearson and Spearman, respectively.}
\label{tab:controltask2}
\end{table*}

\begin{table*}[ht]
\centering
\begin{tabular}{llllllll}
\toprule
\textbf{Model} & \textbf{Task} & \textbf{T2-T} & \textbf{T2-D} & \textbf{T5-A} & \textbf{T5-S} & \textbf{T5-C} & \textbf{T5-Avg} \\
\midrule
BERT-base & control & 0.5173 & 0.5015 & 0.5357 & 0.5036 & 0.5179 & 0.5191 \\
 & original & 0.8286 & 0.9167 & 0.6071 & 0.6179 & 0.5857 & 0.6036 \\
Sheared-llama & control & 0.5133 & 0.5448 & 0.5607 & 0.4750 & 0.5321 & 0.5226 \\
 & original & 0.8235 & 0.9074 & 0.5679 & 0.5750 & 0.5786 & 0.5738 \\
Sheared-llama (Bi + BNTP) & control & 0.5163 & 0.5509 & 0.5143 & 0.5000 & 0.5179 & 0.5107 \\
 & original & 0.7653 & 0.8750 & 0.5607 & 0.5179 & 0.5464 & 0.5417\\
 \bottomrule
\end{tabular}
\caption{This table presents a comparison of accuracy between the control task and the original task in Task 2 and 5. T2-T (Task 2 telicity subtask), T2-D
(Task 2 duration subtask), T5-A/S/C (Task 5 with the absolute difference/cosine similarity/concatenation).}
\label{tab:controltask3}
\end{table*}

\begin{table*}[ht]
\centering
\begin{tabular}{lllllll}
\toprule
\textbf{Model} & \textbf{Subtask} & \textbf{Index1} & \textbf{Index2} & \textbf{Index3} & \textbf{Index4} & \textbf{Index5} \\
\midrule
BERT-base & subject animacy & 0.507 & 1.000 & 0.519 & 0.507 & 0.527 \\
 & verb causative & 0.511 & 0.527 & 1.000 & 0.508 & 0.510 \\
 & verb dynamic & 0.516 & 0.511 & 1.000 & 0.516 & 0.530 \\
 & object animacy & 0.515 & 0.533 & 0.501 & 0.482 & 1.000 \\
Sheared-Llama & subject animacy & 0.482 & 1.000 & 0.491 & 0.482 & 0.499 \\
 & verb causative & 0.511 & 0.532 & 1.000 & 0.511 & 0.518 \\
 & verb dynamic & 0.515 & 0.494 & 1.000 & 0.515 & 0.509 \\
 & object animacy & 0.507 & 0.498 & 0.501 & 0.507 & 1.000 \\
Sheared-Llama (Bi + MNTP) & subject animacy & 0.518 & 1.000 & 0.493 & 0.518 & 0.494 \\
 & verb causative & 0.511 & 0.535 & 1.000 & 0.511 & 0.522 \\
 & verb dynamic & 0.515 & 0.497 & 1.000 & 0.485 & 0.513 \\
 & object animacy & 0.507 & 0.514 & 0.498 & 0.507 & 1.000\\
 \bottomrule
\end{tabular}
\caption{This table presents the accuracy of three semantic features in Task 1 without giving context.}
\label{tab:task1withoutcontext}
\end{table*}

\begin{table*}[ht]
\centering
\begin{tabular}{llllllll}
\toprule
\textbf{Model} & \textbf{Context} & \textbf{Task2-T} & \textbf{Task2-D} & \textbf{Task3- P} & \textbf{T3- S} & \textbf{Task4- P} & \textbf{T4- S} \\
\midrule
BERT-base & with & 0.8286 & 0.9167 & 0.8215 & 0.7971 & 0.6706 & 0.5916 \\
 & without & 0.7480 & 0.8302 & 0.7105 & 0.6874 & 0.5405 & 0.5015 \\
Sheared-Llama & with & 0.8235 & 0.9074 & 0.8219 & 0.7946 & 0.6408 & 0.5785 \\
 & without & 0.7531 & 0.8364 & 0.7270 & 0.7075 & 0.5880 & 0.5403 \\
Sheared-Llama (Bi + MNTP) & with & 0.7653 & 0.8750 & 0.7157 & 0.7079 & 0.5788 & 0.4976 \\
 & without & 0.7367 & 0.8318 & 0.7034 & 0.6830 & 0.5714 & 0.5278\\
\bottomrule
\end{tabular}
\caption{This table presents a comparison of accuracy in Task 2, 3, 4 and 5 with and without giving context. T and D in Task 2 denote Telicity and Duration, and P and S in Task 3 and 4 represent the Pearson and Spearman.}
\label{tab:task2withoutcontext}
\end{table*}



\end{document}